\documentclass[a4paper,twoside]{article}

\usepackage{epsfig}
\usepackage{subcaption}
\usepackage{calc}
\usepackage{amssymb}
\usepackage{amstext}
\usepackage{amsmath}
\usepackage{amsthm}
\usepackage{multicol}
\usepackage{pslatex}
\usepackage{apalike}
\usepackage{array, booktabs, xcolor, arydshln}
\usepackage{SCITEPRESS}     

\begin{document}

\title{GAN-based Intrinsic Exploration For Sample Efficient Reinforcement Learning}
\author{\authorname{Doğay Kamar\sup{1}, Nazım Kemal Üre\sup{1,2} and Gözde Ünal\sup{1,2}}
\affiliation{\sup{1}Faculty of Computer and Informatics, Istanbul Technical University, Istanbul, Turkey}
\affiliation{\sup{2}Artificial Intelligence and Data Science Research Center, Istanbul Technical University, Istanbul, Turkey}
\email{\{kamard, ure, gozde.unal\}@itu.edu.tr}
}


\abstract{In this study, we address the problem of efficient exploration in reinforcement learning. Most common exploration approaches depend on random action selection, however these approaches do not work well in environments with sparse or no rewards. We propose Generative Adversarial Network-based Intrinsic Reward Module that learns the distribution of the observed states and sends an intrinsic reward that is computed as high for states that are out of distribution, in order to lead agent to unexplored states. We evaluate our approach in Super Mario Bros for a no reward setting and in Montezuma's Revenge for a sparse reward setting and show that our approach is indeed capable of exploring efficiently. We discuss a few weaknesses and conclude by discussing future works.}

\onecolumn \maketitle \normalsize \setcounter{footnote}{0} \vfill
\section{\uppercase{Introduction}}
In reinforcement learning, an agent learns which action to take depending on the current state by trying to maximize the reward signal provided by the environment \cite{Sutton1998}. The agent is not given any prior information about the environment or which action to take, but instead, it learns which actions return more reward in a trial-and-error manner. To do so, agents are usually incentivized to explore the state-action space before committing to the known rewards in order to avoid  exploitation of a non-optimal solution.

Most common approaches to exploration, such as $\epsilon$-greedy in Deep-Q-Network (DQN) \cite{dqn}, adding Ornstein–Uhlenbeck to action in Deep Deterministic Policy Gradient (DDPG) \cite{ddpg} or maximizing entropy over the action space in Asynchronous Advantage Actor-Critic (A3C) \cite{a3c}, rely on increasing the probability of a random action. This approach works fine in an environment with dense reward signals. However, in sparse or no reward settings, the agent fails to find a reward signal to guide itself, thus failing to find a solution.

Related works have focused on providing an extra intrinsic reward to the agent in addition to the extrinsic reward, i.e. the reward from the environment. Prediction-based exploration models estimate the novelty of an observation by learning the environment dynamics and assigning higher rewards to the observation which the future of it can not predict well, i.e. it makes the agent act "curiously" \cite{schmid91,icm,bradly}. In another work, environment dynamics is modeled using variational inference and the novelty of a state is measured through prediction using this model \cite{vime}. \cite{pathak19disagreement} proposed a framework with multiple dynamic models, and the intrinsic reward is calculated by the disagreement between the dynamic models. 

Another set of methods are based on count-based exploration, by counting the number of times a state is visited and encouraging the agent to visit less-visited states. However, if the state space is high dimensional, it is not feasible to keep track of all the states. Recent studies proposed density models to approximate how frequently a state is visited \cite{bellemare,ostrovski,zhao}. \cite{tang} benefited from hashing in order to shrink the state-space to keep the count of the visitations to states.

In "Random Network Distillation (RND)", features of an observation are calculated through a fixed, randomly initialized network and another network is trained to predict these features, with the error being the intrinsic reward \cite{RND}. "Never Give Up" adds on to RND by a memory-based episodic reward module, where they keep the embeddings of the states observed in the current episode and compare the new states to the ones in the memory, discouraging the agent to visit the same states in the episode, while also using RND for the long-term exploration \cite{NGU}.  DORA framework creates an identical MDP to the task at hand with E-values fixed to 0 instead of Q-values for every state-action pair, and assigns a bonus intrinsic reward for how well the framework can predict the E-value in the current state \cite{dora}.

Instead of computing an intrinsic reward, \cite{go-explore} proposes "Go-Explore", a method that focuses on exploring until a solution is found before starting policy learning. In "Go-explore", a memory is maintained to keep the visited states, which are encoded into lower-sized cells, and periodically, a state is selected from the memory to go back to and explore from. Selection of the state is probabilistic, however, the probabilities are assigned using a heuristic, which tries to assign higher probabilities to "interesting" states, i.e. states that have better chances to lead to a good exploration. Once a solution is found through exploration, the agent goes through a robustification phase using imitation learning. A similar idea named "Diverse Trajectory-conditioned Self-Imitation Learning (DTSIL)" is proposed by \cite{Guo2019EfficientEW}, where they train a policy using Self-Imitation Learning \cite{SIL} to follow trajectories stored in a memory, which lead agent to less-frequently-visited states, in order to learn a policy for exploring the environment. 

This study follows the trend of providing an intrinsic reward to the agent by using Generative Adversarial Networks (GAN) \cite{gan}. GANs consist of two networks: a generator $G$ tasked with generating synthetic data similar to real data, and a discriminator $D$ tasked with measuring the probability of the input being real to discriminate real and generated data. Both $G$ and $D$ have tasks adversarial to each other as $G$ tries to fool $D$ that the data it generates is real, and $D$ tries to successfully detect data generated by $G$. The convergence of GAN happens when $G$ is able to generate data indistinguishable to $D$, and $D$ has 50\% accuracy of detecting fake data, meaning $D$ starts to randomly guess. $G$'s ability to generate new data comes from its learning and fitting to the distribution of the real data, therefore, in theory, with the right input, $G$ should be able to generate exactly the same sample from the original data distribution. This idea is realized in the anomaly detection task in \cite{anogan} and \cite{fanogan}, in which, after the training of GAN, $G$ is forced to generate the query sample and an anomaly score is assigned by measuring the difference between generated and query sample. A high anomaly score means the query sample is an out-of-distribution sample. Inspired by this work \cite{fanogan}, we propose a GAN-based exploration method for reinforcement learning, where a GAN is trained using the visited state observations. The trained $G$ is used to assign an intrinsic reward to the agent, and unobserved novel states are assigned higher rewards due to them not being in the distribution of the observed states used in GAN's training. 

A recent study has also utilized GANs for the task of exploration \cite{gaex} by using $D$ to distinguish between visited and novel states. However, their approach trains $G$ less frequently so that $D$ does not lose the ability to distinguish between fake and real data. This approach has two weaknesses: (a) $G$ needs to train just good enough to incentivize $D$ to improve itself, but not good enough so that GAN does not reach convergence, and with the right hyperparameter setup, the task of keeping this balance is hard to accomplish; and (b) $D$ is trained to separate real data from what $G$ generates, however, there is no guarantee on how $D$ will act on an unseen instance, resulting in unreliable feedback. In our proposed method, the GAN is trained to convergence so that $G$ is able to generate data similar to real states, and an encoder $E$ is trained to map a newly observed state to the latent noise space, to then be regenerated by $G$. The difference between the observed state and the generated state is then fed as the intrinsic reward to the learning agent. We chose to use GANs for this task because (a) with this approach, learning the dynamics of the environment is not required, we are only interested in learning the distribution of the observed states, and (b) GANs are very widely studied and its use-cases have expanded to many areas. We believe that GANs can be integrated into reinforcement learning and showcasing a successful application will open the gates for the future studies.

We evaluate our proposed method on Montezuma's Revenge, an Atari game that is notoriously hard to explore due to rewards being very sparse and difficult to reach from the initial position by random exploration \cite{bellemare}. In addition to the sparse reward setting, we also evaluate our method in a no-reward setting in the game of Super Mario Bros. \cite{gym-super-mario-bros}. We show by that using only the intrinsic reward, the agent is capable of exploring the environment. We discuss how the method can be further improved and invite readers to build upon the proposed method.

\begin{figure*}
  \centering
  \includegraphics[width=\textwidth]{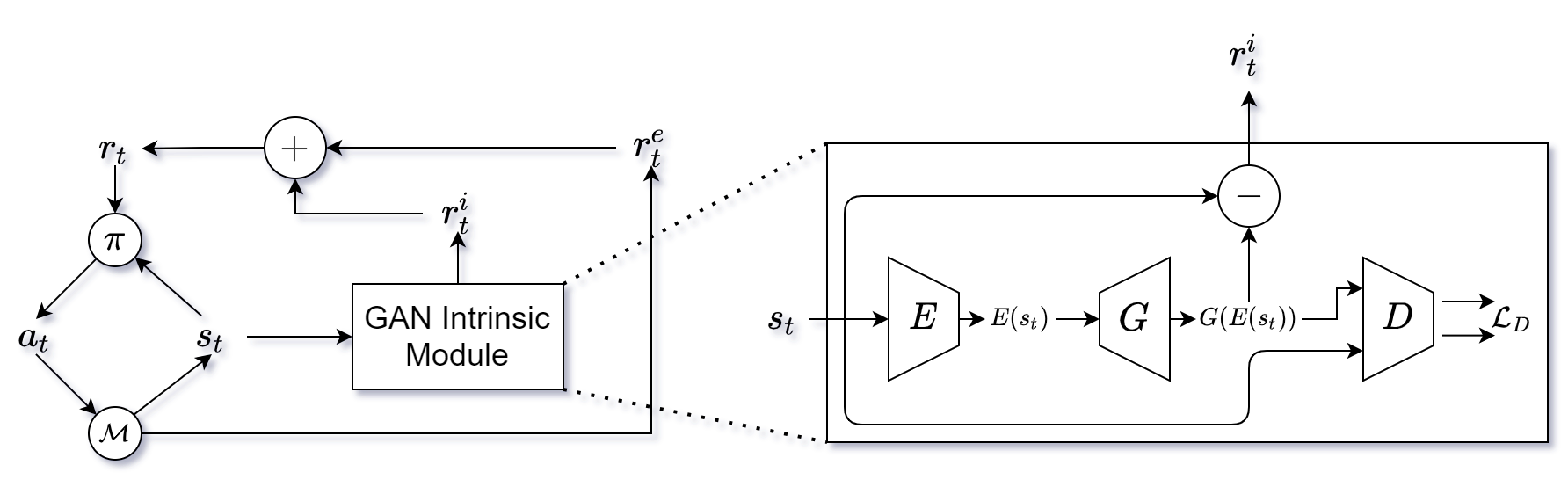}
  \caption{A general reinforcement learning framework with our proposed GAN-based Intrinsic Reward Module (GIRM). At each time step $t$, agent observes a state $s_t$ from the environment MDP $\mathcal{M}$, chooses an action depending on its policy $\pi$, and receives an extrinsic reward $r^{e}_{t}$ from the environment and an intrinsic reward $r^{i}_{t}$ from GIRM. In GIRM, the encoder $E$ maps $s_t$ to the input noise space of the generator $G$ and outputs $E(s_{t})$. $G$ generates a synthetic state $G(E(s_{t}))$ and the error measured between $s_t$ and $G(E(s_{t}))$ is then sent to the agent as an intrinsic reward. Operation $-$ in the GIRM module is not necessarily subtraction, but could be any function that computes the residual between $s_t$ and $G(E(s_{t}))$. $\mathcal{L}_{D}$ is used for training the Encoder.}
  \label{fig:mainfigure}
\end{figure*}

\section{PRELIMINARIES}
\textbf{Generative Adversarial Networks.} Proposed by \cite{gan}, Generative Adversarial Networks (GAN) is a framework that consists of two neural networks, called generator and discriminator, competing against each other in order to train the generator to be able to generate synthetic data similar to real data. Discriminator $D$ is optimized to detect if the input sample is real or generated, and the generator $G$ tries to fool the discriminator by generating data similar to the real data $\boldsymbol{x}$ from an input noise $\boldsymbol{z}$ sampled from a distribution $p_{\boldsymbol{z}}$. $D$ outputs the probability of the input being real data, and is originally optimized to maximize $\log D(\boldsymbol{x})+\log (1-D(G(\boldsymbol{z})))$ while $G$ is optimized to minimize $\log (1-D(G(\boldsymbol{z})))$. In an optimal training, $G$ is able to learn the distribution of the real data and can sample new synthetic data from the same distribution.

\textbf{Reinforcement Learning.} In reinforcement learning, environments are modeled by a Markov Decision Process (MDP), defined as $\mathcal{M} = (\mathcal{S}, \mathcal{A}, \mathcal{R}, \mathcal{P}, \rho_{0})$, where $\mathcal{S}$ is the set of states, $\mathcal{A}$ the set of actions, $\mathcal{R}$ the reward function, $\mathcal{P}$ the state transition probability distribution and $\rho_{0}$ the distribution of the initial state. Given a state $s_t$ in $t^{th}$ time step, the agent takes an action $a_t$ according to its policy $\pi(s_{t})$, receives a reward $r_t$ and observes the next state $s_{t+1}$. The aim is to find a policy $\pi*$ that maximizes the expected total discounted reward,
\begin{equation}\label{eq1}
\pi^{*} = \arg\max _{\pi} \mathbb{E}_{\pi(s_{t}),\mathcal{P}}\left[\sum_{t} \gamma^{t} r_{t}\right]
\end{equation}
where $\gamma \in[0,1)$ is the discount factor that determines the weight of rewards in the distant future compared to present. 

\textbf{Exploration in Reinforcement Learning.} One of the longstanding problems of reinforcement learning is to explore the environment efficiently before agent exploits a sub-optimal but positive reward it receives. The most common approaches for exploration depend on diversifying action selection, meaning that agent ideally performs a random action in a state it did not previously observe, rather than sticking to a single action. These approaches do not necessarily guide agent to unvisited states, but occasionally lets the agent take a random action that might guide it to a better reward. However, if the reward signals in an environment are sparse or nonexistent, the policy $\pi$'s behaviour will be random at any state as the agent does not encounter a reward signal to guide itself. In such problems, an efficient exploration method is necessary to discover unvisited states in order to find the sparse rewards to learn the $\pi^{*}$.

\section{EXPLORATION GUIDED BY GAN-BASED INTRINSIC REWARD MODULE}
\label{section:method}
Influenced by the anomaly detection framework of \cite{fanogan}, we propose a "\textbf{GAN-based Intrinsic Reward Module (GIRM)}" in order to incentivize the policy learning agent to explore unvisited states. To do so, at every time-step, in addition to the extrinsic reward received from the environment $r^{e}_t$, GIRM will also feed the agent with an intrinsic reward $r^{i}_t$, which will be higher for the unexplored states. Policy of the agent is then updated to maximize the total reward, $r_t = r^{e}_t + r^{i}_t$. Providing an intrinsic reward is especially important for the environments with sparse or no rewards, because in such environments, the agent usually fails to find the states with rewards and it needs reinforcement to find the set of actions that will maximize the total reward received. With the intrinsic reward, the agent will now be reinforced into exploring the environment to find the extrinsic rewards. Figure \ref{fig:mainfigure} illustrates the proposed GIRM-enhanced reinforcement learning framework.

The first goal of GIRM is to learn the distribution of the visited states. To that end, we train a GAN with the visited states as the train data, using the improved WGAN architecture \cite{improvedWGAN}. Since the generator $G$ is trained to minimize the divergence between $G$'s sample distribution and the distribution of the training data, $G$ is trained to fit to the distribution of the visited states, i.e. $G$ learns the mapping $\boldsymbol{z} \rightarrow G(\boldsymbol{z})$ where $G(\boldsymbol{z})$ is a synthetic state similar to the visited states. 

After training of the GAN is complete, an encoder $E$ is trained to learn the mapping $s_{t} \rightarrow \boldsymbol{z}$, i.e. $E$ maps the state $s_t$ to the input noise distribution of $G$. The aim is to find the $\boldsymbol{z}$ that is mapped to a $G(\boldsymbol{z})$ which is the most similar synthetic state in the distribution of $G$, to the input state $s_t$. To that aim, first, we define an MSE loss to minimize the difference between the input $s_t$ and the recreated image $G(\boldsymbol{z}) = G(E(s_{t}))$:
\begin{equation}\label{eq2}
\min _{E} \mathcal{L}\left(s_{t}\right)= \frac{1}{n} \|s_{t}-G(E(s_{t}))\|^{2}
\end{equation}
where $n$ is the number of dimensions in the state space (if states are images, n represent the number of pixels). As pointed in \cite{anogan} and \cite{fanogan}, the discriminator $D$ is able to learn the feature representations of the inputs in an intermediate layer, and in addition to the loss in Eq. \ref{eq2}, minimizing the difference between the feature encodings of the training data and the recreated data improves the training of $E$. Therefore, we adapt the loss of feature encodings of $s_t$ and $G(E(s_{t})$ as $\mathcal{L}_{D}\left(s_{t}\right)= \frac{\lambda}{n_{d}} \|f(s_{t})-f(G(E(s_{t})))\|^{2}$ and finalize the loss function for $E$ as:
\begin{equation}\label{eq3}
\begin{split}
\min _{E} \mathcal{L}\left(s_{t}\right) + &\mathcal{L}_{D}\left(s_{t}\right) = \frac{1}{n} \|s_{t}-G(E(s_{t}))\|^{2}+ \\
                                &\frac{\lambda}{n_{d}} \|f(s_{t})-f(G(E(s_{t})))\|^{2}
\end{split}
\end{equation}
where $f$ is the output of an intermediate layer in $D$, $n_d$ number of the dimensions in the feature representations, and $\lambda$ is a hyperparameter scaling the feature representation loss. After $E$ is trained, as illustrated in Fig. \ref{fig:mainfigure}, for an observed state by the agent $s_t$, the intrinsic reward is computed as:

\begin{equation}\label{eq4}
r^{i}_{t} = \frac{1}{n} \|s_{t}-G(E(s_{t}))\|^{2},
\end{equation}
which is the MSE between $s_t$ and $G(E(s_{t})$. If $s_t$ is a frequently visited state, it will belong to the distribution learned by $G$, therefore it can be recreated and $r^{i}_{t}$ will be low. On the contrary, if $s_t$ is an unvisited state, the difference between $s_t$ and $G(E(s_{t})$ will be significant and $r^{i}_{t}$ will be high, reinforcing the agent to explore towards $s_t$. Notice that, we do not multiply $r^{i}_{t}$ with a constant scalar to scale it. It is true that in different environments, the scale of the intrinsic rewards will be different, however as the distribution of the visited states change, so will the scale of intrinsic rewards. Therefore, instead of a constant scalar, we calculate the exponentially weighted moving average($EMA$) and variance($EMV$) of the rewards as: 

\begin{equation}\label{eq5}
EMA =
    \begin{cases}
      r^{i}_{1}, & \text{if}\ t=1 \\
      \alpha*r^{i}_{t} + (1-\alpha)*EMA, & \text{otherwise}
    \end{cases}
\end{equation}

\begin{equation}\label{eq6}
EMV =
    \begin{cases}
      0, & \text{if}\ t=1 \\
      \begin{split}\alpha*(r^{i}_{t}-EMA)^2 + \\ (1-\alpha)*EMV\end{split}, & \text{otherwise}
    \end{cases}
\end{equation}
and then, standardize the intrinsic reward:
\begin{equation}\label{eq7}
r^{i}_{t} = \frac{r^{i}_{t} - EMA}{\sqrt{EMV}}. 
\end{equation}

In this study, we used $\alpha = 0.01$. The standardization ensures that the distribution of the rewards will have the same mean and variance throughout the training and in different environments. In Eq. \ref{eq7}, the intrinsic reward is standardized to have a mean of 0 and a variance of 1, and in this study, we used this equation for standardization, however the equation can be modified to have different mean and variance if needed. With the standardization of $r^{i}_{t}$, the optimization problem in Eq. \ref{eq1} becomes:
\begin{equation}\label{eq8}
\pi^{*} = \arg\max _{\pi} \mathbb{E}_{\pi(s_{t}),\mathcal{P}}\left[\sum_{t} \gamma^{t} (r^{e}_{t}+r^{i}_{t})\right].
\end{equation}

\begin{figure*}
  \centering
  \includegraphics[width=\textwidth]{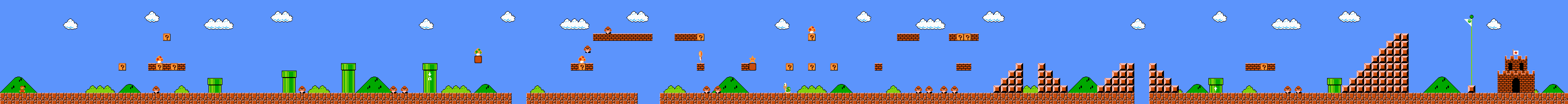}
  \caption{A full view of the first level in the game Super Mario Bros. Player starts at the left side of the level and only perceives the surroundings of the character it is controlling, while the camera is always centered at the character in the horizontal-axis. To complete the level, the player needs to reach the castle at the far right of the level. With no reward given, a reinforcement learning agent only depends on exploration to find the end of the level. }
  \label{fig:mariolevelimage}
\end{figure*}

\begin{figure}
  \centering
  \includegraphics[width=0.45\textwidth]{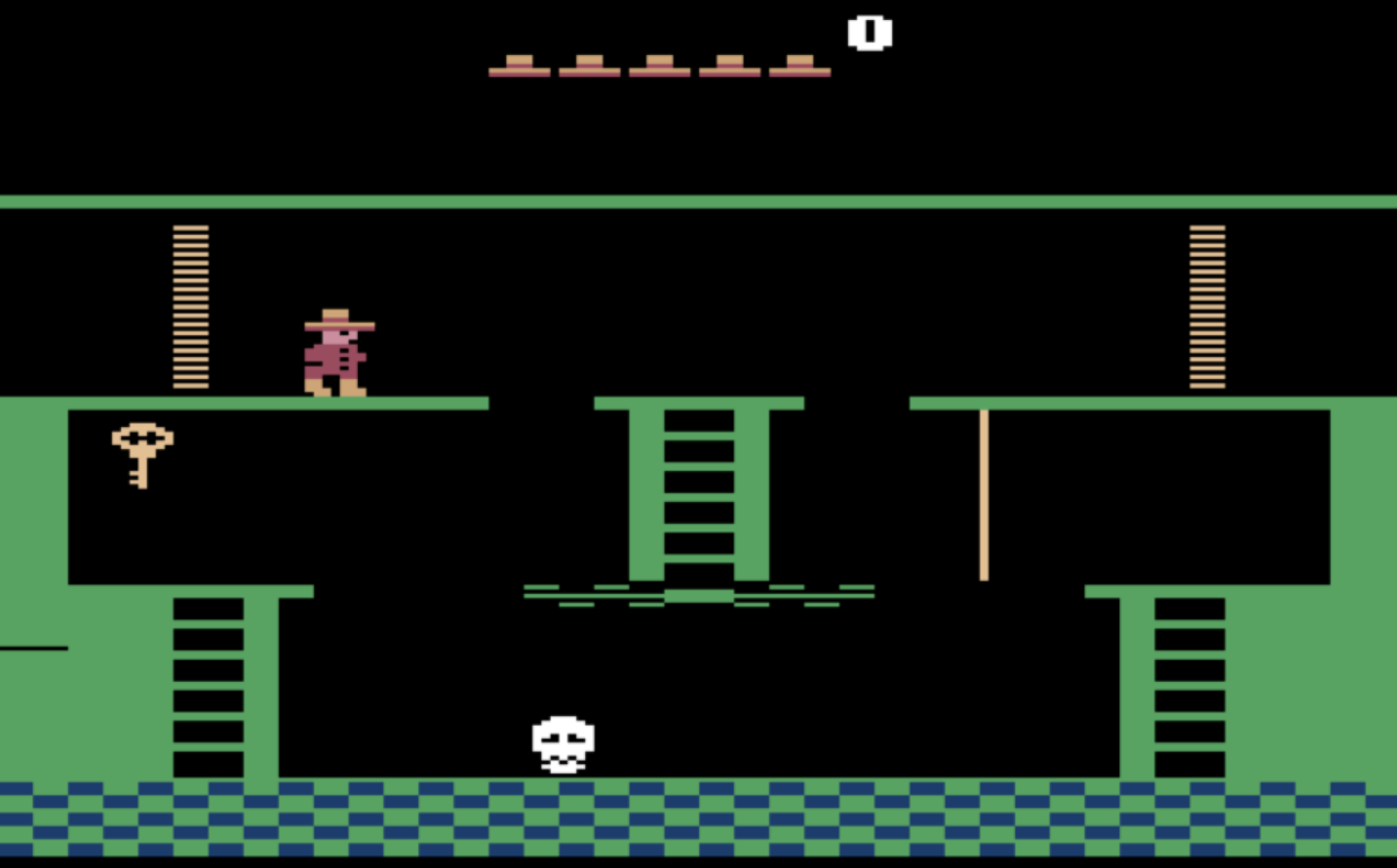}
  \caption{First room in the game Montezuma's Revenge. The player starts the game in this room and needs to collect the key first, and then go to one of the doors to escape from the room. Only rewards are received when a key is grabbed or a door is unlocked. The player needs to complete a complex set of actions to reach the key, as it needs to climb down two stairs, jump to a rope, avoid an enemy, and then climb up a stair while avoiding death. The camera is static, the player has the full view of the room and it only changes when the player goes to another room. }
  \label{fig:montezumalevelimage}
\end{figure}

GIRM is trained periodically during the policy learning. To do so, the states observed by the agent are stored in a memory $M=s_{0}, s_{1},...,s_{N}$ with a pre-defined capacity $N$. As soon as $M$ reaches full capacity, we stop policy learning of the agent, and train GIRM with the data stored in $M$ using stochastic gradient descent. After training of the GIRM is finished, $M$'s content is cleared and the policy learning, as well as storing observed states to $M$, restarts. Every time $M$ reaches full capacity, the GIRM is trained with the new data as the distribution of visited states keeps changing during the policy learning. Since GIRM consists of only random networks before the first time $M$ is full, GIRM does not provide an intrinsic reward until the first training phase, and the agent only receives the extrinsic rewards from the environment. Until the first training phase of GIRM is finished, we use exploration by randomly selected actions. Moreover, the first time $M$ is full, GIRM is trained for larger number of epochs to train it until convergence. Afterwards, GIRM is fine-tuned with the newly observed data for a lesser number of epochs.

\section{EXPERIMENTS}

\begin{table*}[h]
\caption{Comparison of agents in Super Mario Bros. in terms of furthest point they explored and percentage of the states that are further from certain point out of all of the states they visited in horizontal axis. }\label{tab:marioresults} \centering
\begin{tabular}{|c|c|c|c|c|c|}
  \hline
  \textbf{Method} & \textbf{\# of Frames} &\textbf{Furthest Visited} & \textbf{ visited $>$ 500}  & \textbf{visited $>$ 1000} & \textbf{visited $>$ 1500} \\
  \hline
  A2C & 32M & 1673 & 41.3\% & 2.3\%  & 0.1\% \\
  \hline
  A2C+GIRM & 8M & 3150 (End of level) & 60.3\% & 21.5\% & 4.1\%\\
  \hline
\end{tabular}
\end{table*}

\subsection{Experimental Setup}
As the baseline reinforcement learning algorithm, we used Advantage Actor-Critic (A2C), a synchronous version of A3C \cite{a3c} in the experiments. The main difference between A2C and A3C is, while in A3C, there are a number of workers with their own network weights asynchronously working and updating a global network, in A2C, workers send transition synchronously, and share one global network. In our experiments, we used sixteen workers synchronously running their own environments. For both A2C and GIRM, observed state visuals are converted to grayscale from RGB and resized to $84\times 84$. The actor-critic network has 3 convolutional layers, followed by a fully connected layer. First convolutional layer has 32 filters of 8$\times$8 kernel size with stride 4, second convolutional layer has 64 filters of 4$\times$4 kernel size with stride 2, the third convolutional layer has 64 filters of kernel size 3$\times$3 with stride 1 and finally, the fully connected layer has 512 hidden units. Each layer has Leaky ReLU as its activation function with a negative slope of $0.01$. The output of the fully connected layer is then sent to two separate fully connected layers, one of them selecting the action(actor) and the other predicting the value of the state(critic).

As depicted in Figure \ref{fig:mainfigure}, the GIRM consists of 3 networks: generator $G$, discriminator $D$ and encoder $E$. The size of the input noise is 128 for $G$, and it is created as a 2D input, with the size $1\times1$ and 128 channels.  $G$ consists of 5 transposed convolutional layers, with 64 filters in the first 4 layers and 1 filter in the last layer, each filter having a kernel size of $4\times4$. $D$ consists of 6 convolutional layers with 64 filters in the first 5 layers and 1 filter in the last, with kernel sizes of 4, 5, 5, 3, 1, and 5, respectively. The output of the first 5 layers of $D$ is also used as the intermediate feature representation for the training of $E$. Finally, $E$ is formed with 5 convolutional layers with 64 filters in the first 4 layers and 128 filters in the last, which is the size of the input noise. Kernel Sizes are 4, 5, 5, 3, and 5, respectively. All three networks use Batch Normalization \cite{batch_norm} at the end of each layer except the final ones. Hidden layers in $D$ and $E$ are activated with Leaky ReLU with a negative slope of $0.2$, and hidden layers of $G$ uses ReLU as the activation function, except the final layer which uses tanh as its activation function.  

For all the networks, we used ADAM optimizer \cite{adamoptim} to update the weights. For training the actor-critic network, learning rate is $2.5*10^{-4}$ and for the networks in GIRM module, learning rate is $10^{-4}$. $\beta_{1} = 0.9$ and $\beta_{2} = 0.999$ for all the networks.

To evaluate our proposed module, we set up two environments, one of which has sparse reward signals and the other has no reward signal other than the one in the goal state. We evaluate our module by running an agent with and without the module and compare the results. Later on, we compare our results with the other studies in the field.

First of the environments is the game Super Mario Bros. \cite{gym-super-mario-bros}. Fig. \ref{fig:mariolevelimage} shows the full layout of the first level, which is the level our agent is trained in. The agent can only partially observe the environment, as the states include only the surroundings of the character it's controlling. The initial position is at the far left side of the level and the goal position is the flag next to the castle at the far right side. In this environment, we disable all the extrinsic rewards other than the one given for reaching the goal state. Without an extrinsic reward, the agent depends only on its exploration strategy, and with random exploration, it is a significantly hard task to go past the first few obstacles, let alone reach the goal state.

The second environment is the game Montezuma's Revenge, which is notoriously hard to solve for an AI agent due to its sparse rewards. Fig. \ref{fig:montezumalevelimage} shows the room the player starts in. In order to receive the first extrinsic reward signal, the player needs to navigate to the key. However, the agent needs to select a specific set of actions in order to reach the key, as a fall from any height results in death, so the agent needs to navigate the character to jump to the middle platform, climb down the stairs, jump to the rope and to the platform on right, climb down the second stairs, avoid the enemy and finally climb up the stairs. Producing an action set randomly to carry out all these tasks is especially hard. Differing from Super Mario Bros, the observed environment does not change with the movement of the agent and the background stays static until the agent moves to a different room. This provides a different challenge to the GIRM, as the generated image by the generator $G$ will have the same background whether the state is visited or not, until a new room is explored, and $r^{i}_{t}$ calculated in Eq. \ref{eq4} will only have small differences between states. Standardization in Eq. \ref{eq7} is especially important in this type of environments to make the $r^{i}_{t}$ computed for visited and unvisited states have a meaningful difference.

\subsection{Results and Discussion}

\textbf{Super Mario Bros.} Table \ref{tab:marioresults} compares the agents trained with A2C with and without GIRM module, by the distance they covered in horizontal axis in the first level of the game Super Mario Bros. With no reward signal received, the agent only depends on its exploration strategy to find the goal state of the environment. In the case of no GIRM module, a strategy to maximize the entropy of the actions is used \cite{a3c}. Results show that when GIRM is introduced, the agent can find the goal state, whereas without it, the agent can only reach nearly half point of the level and it can only visit there once. 

In Mario, as shown in Fig. \ref{fig:mariolevelimage}, in order to explore the environment, the agent needs to go right consistently. When there is no extrinsic reward, exploration depends only on the observation of new environmental elements and vertical movement does not provide new information and is only useful for overcoming obstacles. Since the intrinsic rewards from the GIRM are higher in the unobserved states that are on the right side of the level, the agent eventually needs to learn to go right in order to explore efficiently. As shown in Table \ref{tab:marioresults}, the visited states by the agent with GIRM that are further from half point of the level amounts to $4.1\%$ of all visited states. Through the observed rewards from GIRM in the first part of the level, the agent learns to go right as it receives higher rewards from the states on the right side of the map and through this learnt policy, the agent manages to find the end of the level without roaming around in the second half of the level. Even though there is no extrinsic reward, through the rewards from the GIRM, our agent learns to explore the environment by learning to move right until the end of the episode, showing that it is capable of efficient exploration. This is an important trait for the agent to have, as, through the usage of the GIRM, we show that the agent can learn how to explore efficiently in the specific environment setting it is in.

\begin{figure}
  \centering
  \includegraphics[width=0.45\textwidth]{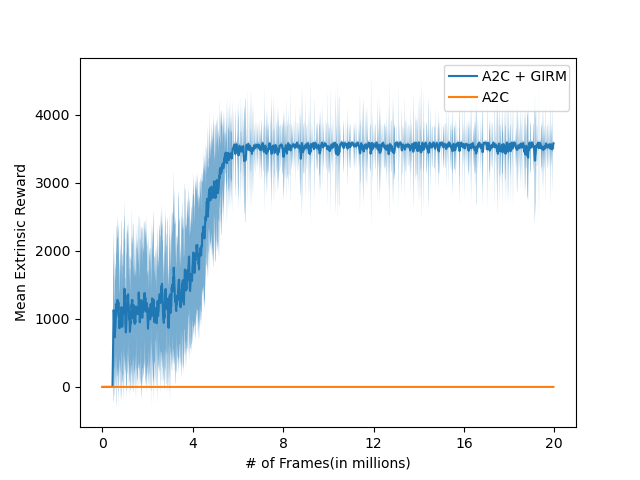}
  \caption{Mean scores during the training of A2C and A2C + GIRM in Montezuma's Revenge. }
  \label{fig:montezumaplot}
\end{figure}

\textbf{Montezuma's Revenge.} An exceptionally hard-to-solve environment, it is very crucial to explore efficiently and find the rewards in Montezuma's Revenge. Our agent trained with the GIRM has shown capable of doing so, guided to rewards and achieved better scores than the baseline A2C agent, as shown in Fig. \ref{fig:montezumaplot}. Agent trained without the GIRM was capable of finding the reward for grabbing key in the first room very few times, and in those cases, it could not find its way out of the room to receive further rewards. Agent trained with the GIRM has shown its capability to further explore the environment, finding rewards in the other rooms as well. However, it should be noted that an early convergence happens as the agent could not break out of the score of 3600. We observed that the agent found a 3000 reward for killing an enemy and started to exploit that big reward rather than exploring. This shows that even though our GIRM guides the agent into exploration, exploiting extrinsic rewards when the difference between intrinsic and extrinsic rewards is high could not be avoided. Results in Table \ref{tab:montezumaresults} further shows that our method is capable of exploring and matching the results of some of the works in the literature but could not beat the performances of the state-of-the-art.

\begin{table}[t]
\caption{Comparison of our results in Montezuma's Revenge against DQN-GAEX \cite{gaex}, Go-explore \cite{go-explore}, NGU \cite{NGU}, RND \cite{RND} and DTSIL \cite{Guo2019EfficientEW}. Scores are reported by the authors of the methods, except the baseline result from A2C and our method, GIRM. Bold indicates the best result. }\label{tab:montezumaresults} \centering
\begin{tabular}{|c|c|c|}
  \hline
  \textbf{Method} & \textbf{\# of Frames} & \textbf{Mean Score} \\
  \hline
  A2C & 20M & 0 \\
  A2C+GIRM & 20M & 3,594 \\
  \hline
  DQN-GAEX & 200M & 420 \\
  Go-explore & ~12B & \textbf{43,791} \\
  NGU & 35B & 10,400 \\
  RND & 16B & 10,070 \\
  DTSIL & 3.2B & 22,616\\
  \hline
\end{tabular}
\end{table}

Another factor affecting the performance of the GIRM is the static background observation in the rooms, resulting in small differences between the observed state and the regenerated state by the generator $G$(even if the observed state is novel) as long as the agent stays in the same room, because the only difference between state observations is the position of the character. We addressed this problem by introducing exponentially weighted moving average ($EMA$) and variance ($EMV$) in Section \ref{section:method} to make even very small differences into meaningful intrinsic rewards. However, when the agent manages to find new rooms in the level, the difference for calculating intrinsic reward grows bigger, which also increases $EMA$ and $EMV$. This results in small differences being not significant enough anymore, hindering the exploration in an already observed room. Nevertheless, as shown by the results in both Super Mario Bros. and Montezuma's Revenge, we can conclude that our proposed GIRM is capable of incentivizing the agent into exploring, heavily outperforming the random exploration approaches that are most commonly used in the literature, however, we also identified a few aforementioned weaknesses as well.

\textbf{Sample Efficiency of GIRM.} As explained, GIRM is trained when the memory $M$ that stores observed states is full and the first training phase of GIRM occurs for a larger number of epochs in order to train it until convergence. This leads to optimized networks in the GIRM even after the first training phase and GIRM becomes capable of computing accurate intrinsic rewards, i.e. rewards that are higher for unobserved states. Because of this, to start leading the agent into exploring the environment efficiently, GIRM needs only the amount of data $M$ initially stored. Our reported results in Table \ref{tab:marioresults} and Table \ref{tab:montezumaresults} uses only 8 million and 20 million of frames for each respective environment. Especially in Montezuma's Revenge, as also shown in Table \ref{tab:montezumaresults}, recent studies have benefited from billions of frames even for efficient exploration. Our method proves to be sample efficient, as it successfully provides the necessary intrinsic reinforcement for the agent with significantly less amount of data.

\section{CONCLUSION AND FUTURE WORK}

Common exploration strategies work well when reward signals are dense, however an additional incentive is needed in environments with sparse or no rewards. In this work, we proposed the "GAN-based Intrinsic Reward Module (GIRM)" to compute an intrinsic reward for a reinforcement learning agent, in order to incentivize the agent to explore the environment it is in. GIRM is trained so that the intrinsic reward is higher for the novel or less-frequently-visited states to guide the agent into exploring the environment. The memory mechanism that we employ in the training of GIRM provides an adaptation to the changing distribution in the observed environment. Since GIRM is an extension module, it can be applied to any reinforcement learning algorithm. In this work, we used Asynchronous Actor-Critic (A2C) as the baseline algorithm, comparing agents trained with and without GIRM.

We show GIRM's capability of exploration in a sparse reward and a no reward setting. We used Super Mario Bros. \cite{gym-super-mario-bros} for the no reward setting and show that the agent trained with GIRM provides the necessary intrinsic reward so that the agent explores and completes the level and outperforms the agent without GIRM. Our agent does so by learning the pattern of going right in order to find unexplored states. For the sparse reward setting, we evaluated our agent in Montezuma's Revenge, an atari game that's been recently used for benchmarking in reinforcement learning, due to its difficulty. While the agent trained without GIRM is not capable of escaping the initial room, with the addition of GIRM, our agent explores multiple rooms throughout the environment, achieving a mean score of 3954. On the one hand, we show that GIRM provides a more efficient exploration strategy, but on the other hand, we observe that our agent converges early, beginning to exploit the high rewards from the environment. 

We also identify another weakness of GIRM through Montezuma's Revenge: standardizing rewards through the usage of $EMA$ and $EMV$ turns small differences between the observed novel state and regenerated state into meaningful intrinsic rewards, however, as the agent begins to explore new rooms, the already high difference between regenerated and novel states gets higher, which also increases the distribution $EMA$ and $EMV$ represents, therefore GIRM loses the capability of assigning meaningful rewards to novel states in the frequently visited rooms. In the future, we would like to address this problem. A potential solution could be leaving out the very high or very low intrinsic reward when updating $EMA$ and $EMV$, treating them as an anomaly. Furthermore, a more efficient reward scaling method could be investigated. 

Another future direction is to make use of GANs to train a model to learn the dynamics of the environment, instead of the distribution of observations. Since the dynamics throughout the environment do not change drastically, a model that learns the dynamics might have a better generalization property throughout the environment. This idea is not a direct improvement to GIRM, but instead, we recommend an idea to utilize GANs in the efficient exploration problem in reinforcement learning with a different approach.

\section*{ACKNOWLEDGEMENTS}

This work is supported by Istanbul Technical University BAP Grant NO: MOA-2019-42321.

\bibliographystyle{apalike}
{\small
\bibliography{references}}

\end{document}